\documentclass[10pt,twocolumn,letterpaper]{article}

\usepackage{iccv}
\usepackage{times}
\usepackage{epsfig}
\usepackage{graphicx}
\usepackage{amsmath}
\usepackage{amssymb}
\newcommand{\mat}[1]{\mathbf{#1}}

\usepackage[pagebackref=true,breaklinks=true,letterpaper=true,colorlinks,bookmarks=false]{hyperref}

\DeclareMathOperator*{\argmax}{arg\,max}

 \iccvfinalcopy 

\def\iccvPaperID{346} 

\ificcvfinal\fi
\begin{document}

\title{Video Fill In the Blank using LR/RL LSTMs with Spatial-Temporal Attentions}

\author{Amir Mazaheri$^\dagger$, Dong Zhang$^\dagger$, and Mubarak Shah$^\dagger$\\
$^\dagger$Center for Research in Computer Vision, University of Central Florida, Orlando, FL 32816\\
{\tt\small amirmazaheri@cs.ucf.edu, dzhang@cs.ucf.edu, shah@crcv.ucf.edu}
}

\maketitle

\begin{abstract}
   Given a video and a description sentence with one missing word (we call it the ``\textbf{source sentence}"), Video-Fill-In-the-Blank (VFIB) problem is to find the missing word automatically. The contextual information of the sentence,  as well as visual cues from the video, are important to infer the missing word accurately. Since the source sentence is broken into two fragments: the sentence's left fragment (before the blank) and the sentence's right fragment (after the blank), traditional Recurrent Neural Networks cannot encode this structure accurately because of many possible variations of the missing word in terms of the location and type of the word in the source sentence. For example, a missing word can be the first word or be in the middle of the sentence and it can be a verb or an adjective. In this paper, we propose a framework to tackle the textual encoding: Two separate LSTMs (the LR and RL LSTMs) are employed to encode the left and right sentence fragments and   a novel structure is introduced to combine each fragment with an {\em external memory} corresponding the opposite fragments. For the visual encoding, end-to-end spatial and temporal attention models are employed to select discriminative visual representations to find the missing word. In the experiments, we demonstrate the superior performance of the proposed method on challenging VFIB problem. Furthermore, we introduce an extended and more generalized version of VFIB, which is not limited to a single blank. Our experiments indicate the generalization capability of our method in dealing with such more realistic scenarios.
\end{abstract}

\section{Introduction \& Related Works}
In computer vision, due to Deep Convolutional Neural Networks (CNNs) ~\cite{krizhevsky2012imagenet,Simonyan2014,szegedy2015going, he2015deep} dramatic success  has been achieved in detection (e.g. object detection ~\cite{deng2009imagenet}) and classification (e.g. action classification  ~\cite{soomro2012ucf101}). Likewise,  Recurrent Neural Networks (RNN) ~\cite{schuster1997bidirectional,hochreiter1997long, chung2014empirical} have been demonstrated to be very useful in Natural Language Processing (NLP) for language  translation. Recently, new problems such as Visual Captioning (VC) ~\cite{Xu2015,Venugopalan2014, vinyals2015show} and Visual Question Answering (VQA)~\cite{Antol2015,Ren2015, malinowski2014multi, agrawal2015vqa,xiong2016dynamic, zhang2015yin} have drawn a lot of interest, as these are very challenging problems and extremely valuable for both computer vision and natural language processing. Both Visual Captioning and Visual Question Answering are related to the Video-Fill-in-the-Blank (VFIB) problem, which is addressed in this paper. 
\begin{figure}
\begin{center}
   \includegraphics[width=1 \linewidth]{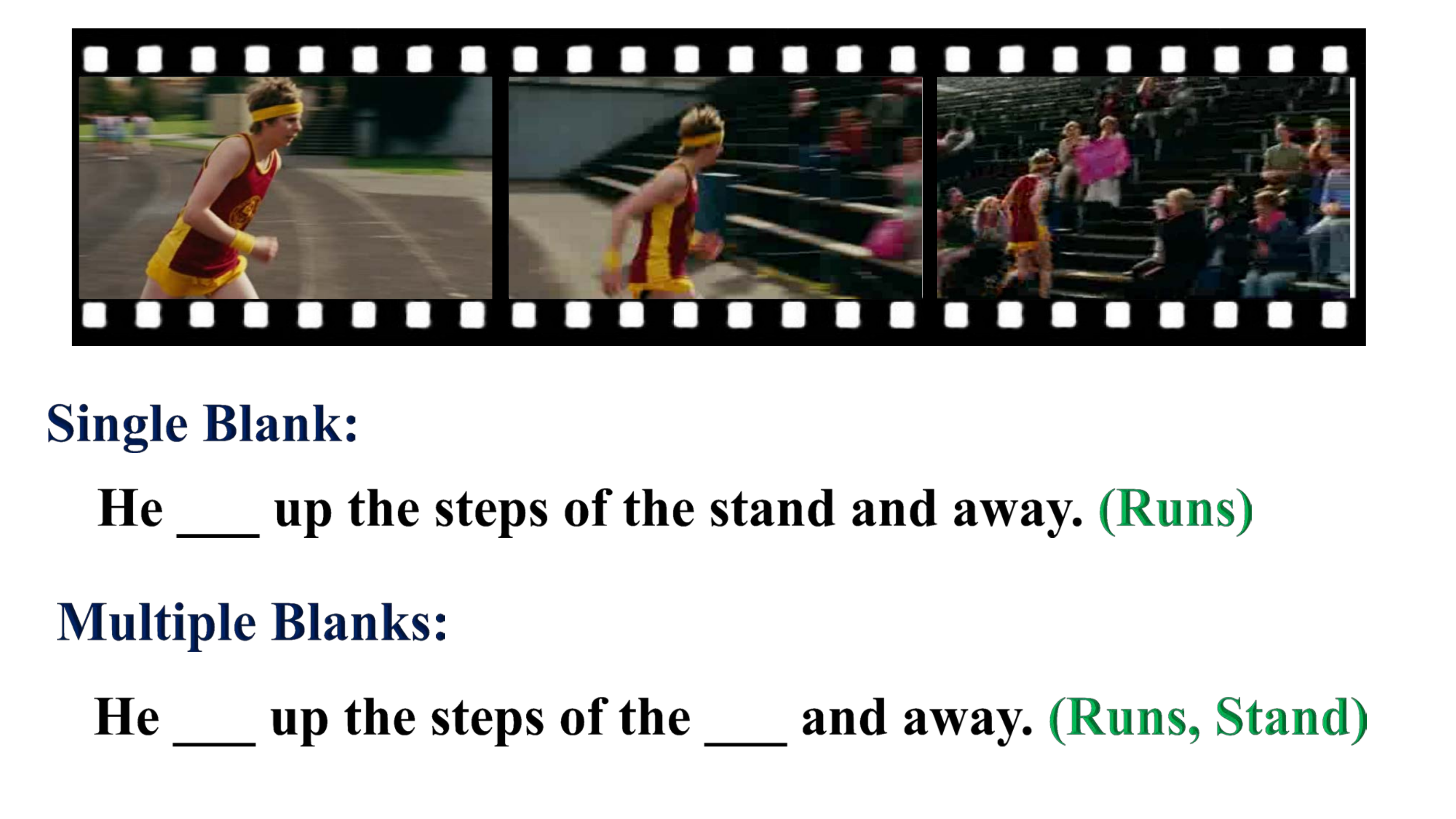}
\end{center}
   \caption{Two examples of the Video-Fill-in-the-Blank problem\cite{rohrbach15cvpr} with a single blank and multiple blanks.}
\label{fig_motivation}
\end{figure}

\begin{figure*}[htb]
\begin{center}
   \includegraphics[width=6.5in]{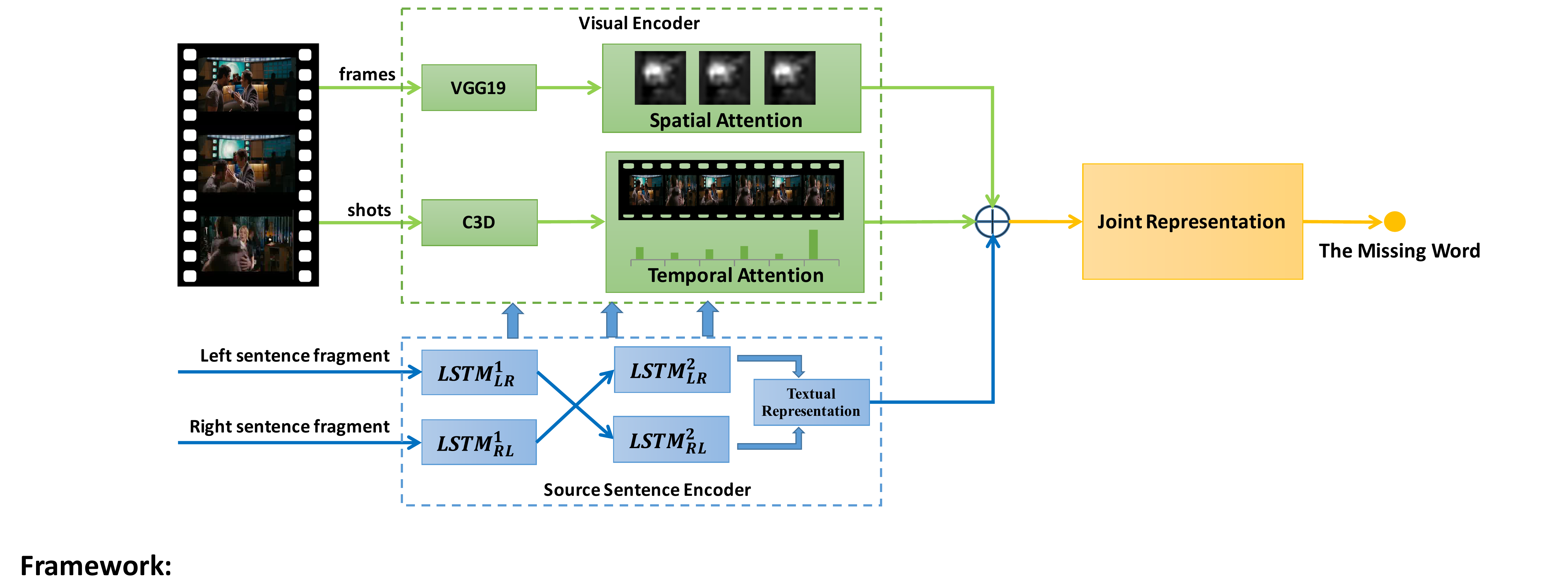}
\end{center}
   \caption{Our proposed method to solve Video Fill In the Blank (VFIB) problem. Source sentence encoding, spatial and temporal attention models are shown.}
\label{fig_framework}
\end{figure*}

Visual Captioning (VC) needs deep understanding of both visual(i.e. image or video) and textual cues.  Recently, several approaches for VC have been introduced. Some of these algorithms focus on leveraging RNNs which take visual input, focus on different regions of the image and leverage attention models to produce captions ~\cite{vinyals2015show,johnson2015densecap}. Furthermore, a dense captioning method is proposed in~\cite{johnson2015densecap}, where for a given image, multiple captions are generated and for each caption a corresponding bounding box is produced. Visual Question Answering (VQA) has deep roots in textual question answering ~\cite{bordes2015large,kumar2015ask,weston2014memory}. The goal of textual question answering is to answer a question based on a collection of sentences. 

The major difference between VQA and VC problems is that in VQA there is a ``question'' as a sentence in addition to the image or video. This makes the problem harder since the question can be about details of the image or video; However, for the VC problem, any factual sentence about the image can be a correct caption. Some methods use a combination of the question and the image/video features through LSTMs ~\cite{malinowski2015ask} and output the answer to the question. Some other methods combine the image features with the words one by one, for instance, the methods in ~\cite{xiong2016dynamic,kumar2015ask}(~\cite{xiong2016dynamic} is an extension of ~\cite{kumar2015ask}), use visual and textual information as input to the dynamic memory networks and convert it to a sequence of words through an RNN with an iterative attention process. In ~\cite{zhang2015yin}, a binary question answering (yes/no) on abstract (unreal) scenes is presented. Another binary answering problem which verifies existence of a relation between two concepts (for example: dogs eat ice cream) is proposed in ~\cite{sadeghi2015viske}, which  uses CNNs to verify the relation. MovieQA~\cite{MovieQA}, presents another form of VQA, however the input in this case is a video and its subtitle. The authors in ~\cite{MovieQA} use  multiple choice questions and demonstrate that the visual information does not contribute to the final results.

Video Fill-In-the-Blank (VFIB)~\cite{maharaj2016dataset} is a relatively new problem and has a broad variety of real world applications such as visual guided report generation, corrupted data recovery, officer report automation for police departments, etc. VFIB is somehow related to Video Question Answering (VQA) problem, however, it has some significant differences and is a more challenging problem. VQA datasets usually have a bias to some specific forms of questions such as location, color, counting, etc. Therefore, the answers for each of these questions are limited to a small dictionary of words. For example, in DAQUAR question answering dataset~\cite{malinowski2014multi}, in many cases, the word ``table'' is the answer to ``What is'' question type, and ``White'' to ``What color'' questions~\cite{Ren2015}. On the other hand, in VFIB problem, there is no limit on the type of the missing word and it can be a verb, adverb, adjective, etc. Furthermore, for VQA problems, there is always a complete sentence to be the ``question''. In this scenario, it is easier to encode the ``question'' in the model (for example, using standard models such as Recurrent Neural Networks) and then use it to predict the answer. Therefore, it is easy and straightforward to use off-the-shelf techniques; However, there is no ``question" as a whole sentence for the VFIB problem, so it is tricky to encode the source sentence using the standard encoding techniques. Also, the blank can be at any location in the sentence, namely, in the middle or very first or last word of the source sentence. Last but not least, for VQA problem, it is very expensive to collect datasets, hence limiting its practical applications. It is time-consuming since the human annotators have to generate questions and answers one by one while watching the videos. Some efforts have been made to make the question-answer generation automatic~\cite{Ren2015}, but these approaches generate a lot of abnormal questions, and they tend to work well only on object related questions.

\begin{figure*}[htb]
\begin{center}
   \includegraphics[width=1\textwidth]{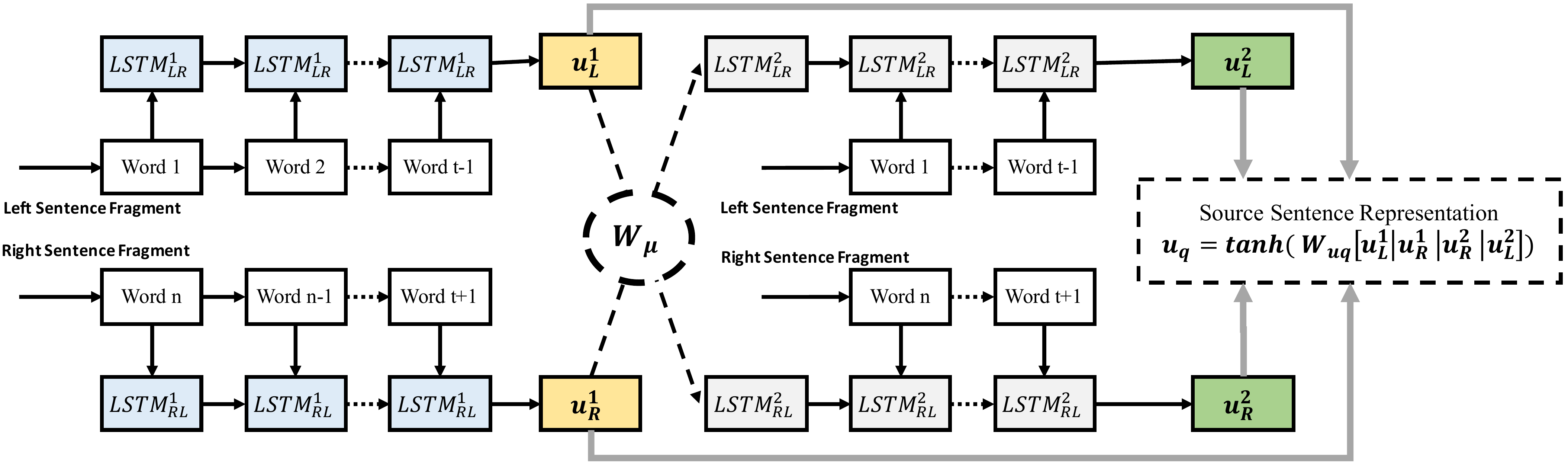}
\end{center}
   \caption{An illustration of the source sentence encoding approach. In the first stage, Each source sentence is formatted as two fragments, the left and right sentence fragments. Each sentence fragment is passed through an LSTM. In the second stage, left and right fragments' LSTMs are combined with a memory from the opposite side.}
\label{fig_textual_encoder}
\end{figure*}

Due to lack of a complete sentence and the various types of missing words and also different possible locations of the missing word in the source sentence, the VFIB problem cannot be well-solved by traditional Recurrent Neural Network (e.g. LSTM). Therefore, we propose a new framework which takes an advantage of the source sentence structure and integrates the spatial and temporal attention modules to fully exploit the visual information. 

This paper makes the following contributions:
First, we propose a novel approach which can encode left and right sentence fragments. It encodes each fragment with a separate LSTM and feeds the output of each fragment to the opposite fragment (for example, from left fragment to the right fragment) as an external memory. This is very different from basic BiLSTM approaches~\cite{berglund2015bidirectional,schuster1997bidirectional}, where each of left and right fragments are encoded separately and then joined by a simple fusion technique (e.g. concatenation, summation, etc). Moreover, we show our text encoding achieves superior performance over other methods. Second, we propose a novel framework for the spatial and temporal attention models, which are trained end-to-end. Third, different visual features are employed by the spatial and temporal attention models, which exploit the visual cues from different domains. Fourth, in our formulation we perform multiple features fusion in end-to-end fashion without any need for pre-training single modules like~\cite{ngiam2011multimodal,nam2016learning}, which deal with multi-modal problems. We also introduce a more general version of VFIB problem and employ source sentences with multiple blanks. We show that our method produces the superior results. 

The organization of the rest of the paper is as follows: In Section \ref{sec_method} we introduce different components of the proposed method in detail; in Section \ref{sec_experiments}, we show experimental results on both single blank and multiple blanks problems; and we conclude in Section \ref{sec_conclusion}.

\label{sec_intro_related}
\section{Method}
\label{sec_method}
\newcommand{\DZ}[1]{{\color{red} DZ: #1}}

In the proposed method, we formulate the VFIB problem as a word prediction problem with two sentence fragments (left sentence fragment ``$q_l$'', and right sentence fragment ``$q_r$'') and the video $\mat{\upsilon}$:
\begin{equation} \label{eq:theory}
    \hat{b} = \argmax_{b \in \beta}{p(b | \mat{q}_l, \mat{q}_r, \mat{\upsilon}, \mat{\theta})},
\end{equation}
where $\beta \subset V$ is the set of words to fill in the blank and $V$ is the dictionary of all the words in our dataset. $\hat{b}$ is the prediction (the word to be filled in the blank), and $\theta$ is the model parameters. Our framework  is depicted in Figure~\ref{fig_framework}. The proposed approach consists of four components: 1) source sentence encoding, 2) spatial attention model, 3) temporal attention model, and 4) inference of the missing word. We discuss these four components one by one in the following subsections. 

\subsection{Source Sentence Encoding} \label{seq:Qemb}

In this section, we introduce how to encode the source sentence which consists of a broken form of left and right fragments and a blank between them. Figure \ref{fig_textual_encoder} shows an illustration of the proposed source sentence encoding approach. The blank can be anywhere in a source sentence (including at the beginning or the end), thus a single LSTM architecture will not work very well, since left or right fragment can be very short and in many cases the missing word depends on both fragments, however, a simple LSTM will fail if one fragment be very short or both fragments be needed for prediction. Also, the blank can be any class of words (e.g. verb, adjective, etc.). These complexities of the source sentence make the textual encoding difficult.

Generally, we treat the source sentence as two fragments; the left fragment from the first word to the word before the blank and the right fragment backward from the last word to the word after the blank. Our source sentence encoding module has three stages. In the first stage, we encode each of the left and right fragments separately with two independent LSTMs. In the second stage, we encode left and right fragments along with the encoded fragments from the opposite side in stage one. Namely, we use the encoded left fragment in the first stage as an external memory to encode the right fragment in stage two and vice versa. We call it ``external memory'', since it is computed using the opposite fragment. In fact, the external memory makes the model to understand each fragment better, since it has some information from the opposite fragment. Finally, the model learns to combine the output of both stages and generates the final source sentence's representation named $u_q$ (Figure~\ref{fig_textual_encoder}). Our approach has two major differences compared to BiLSTM~\cite{berglund2015bidirectional,schuster1997bidirectional}. First, we use the the opposite fragments as an external memory for each sides, and second, our method learns how to combine the left and right encoded fragments.  In the following, we provide more details.

Assume that the source sentence has $n$ words, and the $t$'th word is missing. The left and right fragments' sequences can be embedded as:
\begin{equation}
    \begin{gathered}
        \mat{q}^{1}_{l} = \mat{W}^{1}_{x}  [\mat{x}^1,\mat{x}^2,...,\mat{x}^{t-1}]\\
        \mat{q}^{1}_{r} = \mat{W}^{1}_{x}  [\mat{x}^n,\mat{x}^{n-1},...,\mat{x}^{t+1}],
    \end{gathered}
\end{equation}
where $\mat{x}^i \in {\{0,1\}}^{|V|}$ is an one-hot vector representation of $i$'th word in the source sentence, and $\mat{W}^{1}_{x} \in \mathbb{R}^{ c \times |V|}$ is a word embedding matrix ($|V|$ is the size of the dictionary, and $c$ is the encoding dimension). $\mat{q}^{1}_{l}$ and $\mat{q}^{1}_{r}$ are two sequences where each element ${q}^{1}_{l_{(j)}} \in \mathbb{R}^{c}$ is a continuous vector representing $j$'th word in ${q}^{1}_{l}$ sequence. We model the left and right sentence's fragments separately using two LSTMs:

\begin{equation}
\label{eq_textual_lstms}
\begin{array}{l}
    \mat{u}^{1}_{l} = LSTM^{1}_{LR}(\mat{q}^{1}_{l_{(i)}}), \quad (i=1,...,(t-1))\\
    \mat{u}^{1}_{r} = LSTM^{1}_{RL}(\mat{q}^{1}_{r_{(i)}}), \quad (i=1,...,(n-t))
\end{array}
\end{equation}
where $\mat{u}^{1}_{l}, \mat{u}^{1}_{r} \in \mathbb{R}^{h}$ are the last hidden states from the ``LR'' and ``RL'' LSTMs respectively (Fig.~\ref{fig_textual_encoder}) and $h$ is the LSTMs' hidden state size. Since the missing word is related to both left and right fragments of a sentence, we have an extra stage to encode Left/Right fragments with respect to an external memory coming from the opposite side, namely Right/Left fragments. In this way, the first stage processes each of fragments separately and the second stage processes them with respect to opposite side's encoded representation.
\begin{equation}
    \begin{gathered}
        \mat{q}^{2}_{l} = [\mat{\mu}_l,\mat{W}^{2}_{x}  [\mat{x}^1,\mat{x}^2,...,\mat{x}^{t-1}],\mat{\mu}_l]\\
        \mat{q}^{2}_{r} = [\mat{\mu}_r,\mat{W}^{2}_{x}  [\mat{x}^n,\mat{x}^{n-1},...,\mat{x}^{t+1}],\mat{\mu}_r],
    \end{gathered}
\end{equation}
where $\mat{W}^{2}_{x}$ is of the same size as $\mat{W}^{1}_{x}$ and $\mat{\mu}_r, \mat{\mu}_l \in \mathbb{R}^{c}$ are two external memory vectors obtained by:

\begin{equation}
    \begin{gathered}
    \mat{\mu}_{r} = \mat{u}^{1}_{l}\mat{W}_{\mu}\\
    \mat{\mu}_{l} = \mat{u}^{1}_{r}\mat{W}_{\mu}
    \end{gathered}
\end{equation}
where $\mat{W}_{\mu} \in \mathbb{R}^{h \times c}$ encodes the LSTMs outputs to memory vectors. $\mat{q}^{2}_{l}$ and $\mat{q}^{2}_{r}$ are two sequences while $|\mat{q}^{2}_{l}| - |\mat{q}^{1}_{l}| = |\mat{q}^{2}_{r}| - |\mat{q}^{1}_{r}| = 2$ because of the external memory vectors attached to them and each element of them is a continuous vector. Similar to Eq.~\ref{eq_textual_lstms}, we encode these two sequences with two different LSTMs, namely $LSTM^{2}_{LR}$ and $LSTM^{2}_{RL}$ (Fig~\ref{fig_textual_encoder}), to obtain $\mat{u}^{2}_{l}, \mat{u}^{2}_{r} \in \mathbb{R}^{h}$ as encoded left and right fragments in the second stage. Note that; since $\mat{W}^{1}_{x}$ and $\mat{W}^{2}_{x}$ are two different matrices, LR/RL LSTMs of first  and second stages, observe completely different sequence of vectors. Also, none of these four LSTMs share any parameters with the other ones.

Finally, a proper combination of $\mat{u}^{1}_{l}$, $\mat{u}^{1}_{r}$, $\mat{u}^{2}_{l}$ and $\mat{u}^{2}_{r}$ as the final representation of the source sentence is needed. We concatenate and combine them by a fully connected layer as the final representation of the source sentence:
\begin{equation} \label{eq:qout}
    \mat{u}_q = \tanh(\mat{W}_{uq} [\mat{u}^{1}_{l} | \mat{u}^{1}_{r} | \mat{u}^{2}_{l} | \mat{u}^{2}_{r}]),
\end{equation}
where $\mat{W}_{uq} \in \mathbb{R}^{d \times 4h}$ is a trainable weights matrix applied to learn a proper combination of four vectors. We refer $\mat{u}_{q} \in \mathbb{R}^{d} $ as the source sentence representation (textual feature) in the following sections and it is a bounded vector due to $\tanh(\cdot)$ activation function.

\subsection{Spatial Attention Model} \label{SRep}

\begin{figure}
\begin{center}
   \includegraphics[width=\linewidth]{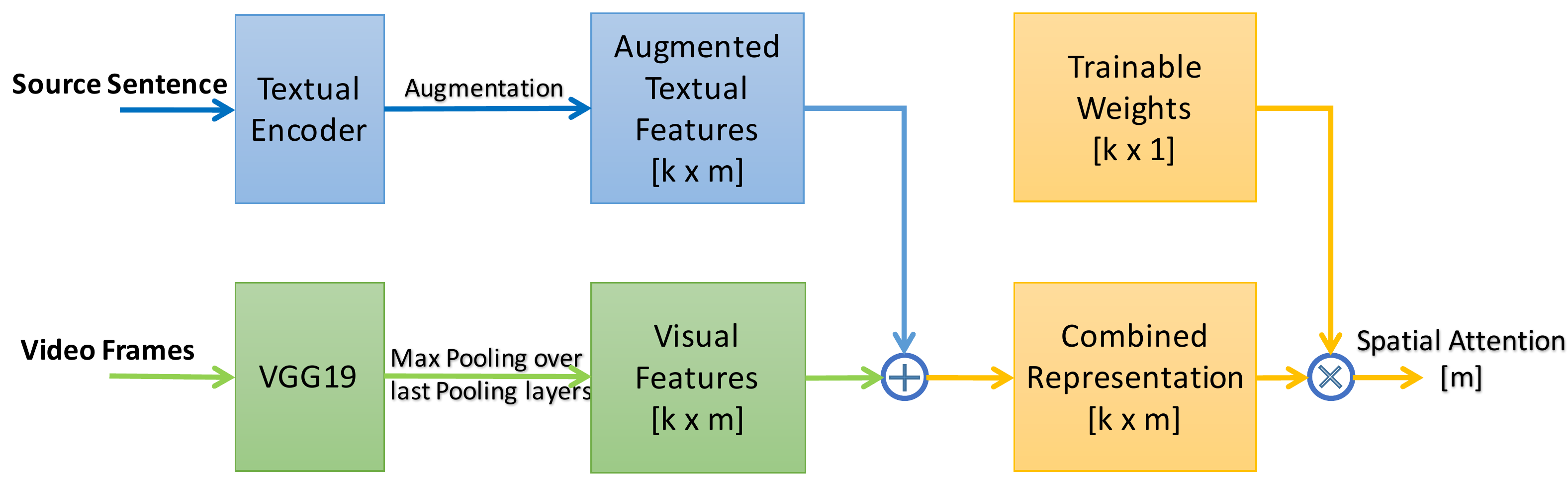}
\end{center}
   \caption{An illustration of the spatial attention model. The model  assigns importance score to each region of an image based on the source sentence.}
\label{fig_spatial_attention}
\end{figure}

We use a CNN (i.e., VGG-19~\cite{Simonyan2014}, more details are provided in Section \ref{featExtraction}) to extract visual features. The output from the last pooling layer of CNNs can be considered as a feature map with spatial information about the input image. Our proposed method learns a spatial attention model~\cite{Yang2015} to use the named spatial feature map. Figure \ref{fig_spatial_attention} shows an illustration of the spatial attention model. First, we apply max-pooling over the raw CNN features (i.e the output from the last pooling layer of VGG-19 pre-trained network) from all video key-frames to obtain a spatial visual feature from the whole video:
\begin{equation} \label{eq:spatialfeatures}
    \mat{\Phi}_F = \tanh(\mat{W}_{f} \Theta(\mat{\Phi}_f(f^t))|_{f^t \in F}),
\end{equation}
where $\mat{\Phi}_f(\cdot)$ is the spatial visual feature map extraction function (more details are provided in Section \ref{featExtraction}), $F$ represents all the video key-frames in $\upsilon$, $f_t$ is a video frame at time $t$, $\Theta(\cdot)$ is the max-pooling function, $\mat{W}_f$ is a trainable transformation matrix,  and $\mat{\Phi}_F \in \mathbb{R}^{d \times m}$ is the intermediate visual feature matrix where each column is a feature vector corresponding to a spatial region in the original video frames, $d$ is the same as $\mat{u}_{q}$ in Eq.\ref{eq:qout}, and $m$ is the number of spatial regions in the video frame.

We use spatial attention model to pool the intermediate visual features $\mat{\Phi}_F$. The first step is to combine the source sentence representation $\mat{u}_q$ with the intermediate visual features $\mat{\Phi}_F$:
\begin{equation} \label{eq:SpatialH}
    \mat{\Psi}_F = \tanh((\mat{W}_F \mat{\Phi}_F) \oplus (\mat{W}_{u} \mat{u}_q + \mat{b}_{u})),
\end{equation}
where $\mat{W}_{F} \in \mathbb{R}^{k \times d}$ and $\mat{W}_{u} \in \mathbb{R}^{k \times d}$ are two transformations on the intermediate visual features and source sentence representation to  align them and have the same dimension ~\cite{Yang2015}. $\mat{b}_{u}$ is the bias term, and $\oplus$ is a summation between a matrix and an augmented vector (i.e. the source sentence representation $\mat{u}_q$ has to be expanded, or repeated $m$ times, in order to have the same dimension as $\mat{W}_F \mat{\Phi}_F$. The matrix $\mat{\Psi}_F \in \mathbb{R}^{k \times m}$ is used to find the final attention scores over all the regions:
\begin{equation} \label{eq:SpatialAttentionVector}
 \mat{p}_{sp} = \textrm{softmax}(\mat{\Psi}_{F}^{T} \mat{w}_{sp}),
\end{equation}
where $\mat{w}_{sp} \in \mathbb{R}^{k \times 1}$ is a trainable weight vector and $\mat{p}_{sp} \in \mathbb{R}^{m \times 1}$ is the spatial attention vector. The final spatial pooled visual vector is a weighted average over all $m$ regional intermediate spatial feature vectors:
\begin{equation} \label{eq:spout}
\mat{u}_{sp} = \Phi_{F} \mat{p}_{sp},
\end{equation}
where $\mat{u}_{sp} \in \mathbb{R}^{d}$ is the spatial pooled visual vector. It is a bounded vector since $\mat{\Phi}_{F}$ is bounded.

\subsection{Temporal Attention Model} \label{TeSe}

\begin{figure}
\begin{center}
   \includegraphics[width=1 \linewidth]{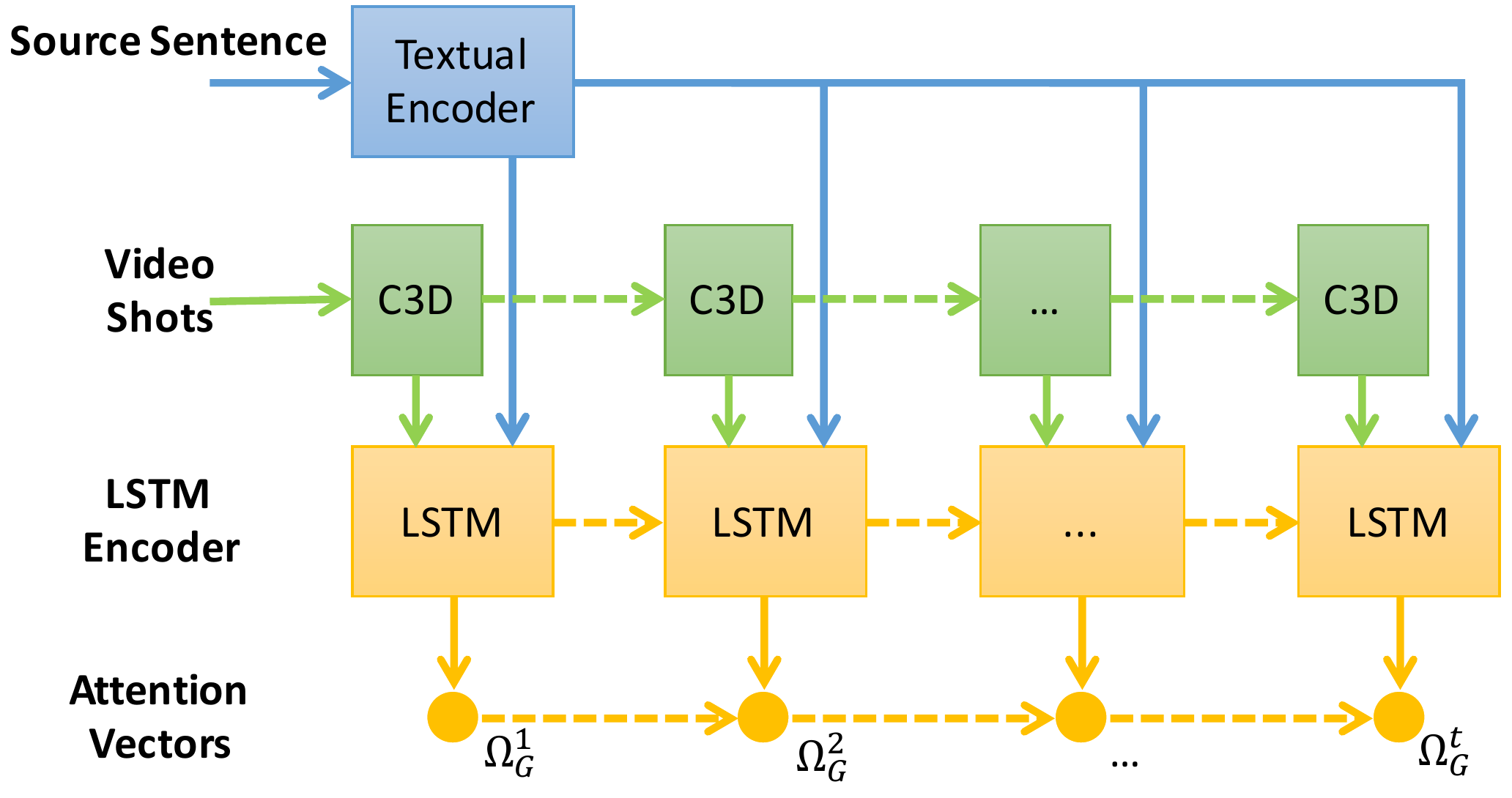}
\end{center}
   \caption{An illustration of the temporal attention model. An LSTM is used to find relevant shots based on the source sentence.}
\label{fig_temporal_attention}
\end{figure}

We  presented the source sentence representation and spatial attention model in previous subsections. Temporal dynamics of a video and the motion information plays a significant role in video understanding, especially in actions and events understanding. However, in our spatial-pooled visual representation, the whole video is represented as one vector, and there is no temporal information. Therefore, we propose to model the temporal dynamics of the video using a temporal attention model. Figure \ref{fig_temporal_attention} shows an illustration of  this component in our approach. We divide a video into a number of shots and represent the shots as below:
\begin{equation} \label{eq:upsilon}
    \mat{\Phi}_G = \tanh(\mat{W}_g [\mat{\Phi}_g(g^1), \mat{\Phi}_g(g^2),..., \mat{\Phi}_g(g^{|G|})]),
\end{equation}
where $\mat{\Phi}_g(\cdot)$ is the feature extraction function (C3D ~\cite{tran2015learning}, which encodes the temporal information. More details are provided in Section~\ref{featExtraction}), $G$ represents the set of video shots and $g^i$ is the $i$th video shot. $\mat{W}_g \in \mathbb{R}^{d \times z}$ is a transformation matrix, where $z$ is the original dimension of the feature vector $\mat{\Phi}_g(g^i)$, and $k$ is the encoding dimension.
We combine the video shots representation $\mat{\Phi}_G \in \mathbb{R}^{d \times |G|}$ and the source sentence representation $\mat{u}_{q}$ as:
\begin{equation}
      \mat{\Psi}_{G} = \tanh((\mat{W}_{G} \mat{\Phi}_G) \oplus (\mat{W}_{u} \mat{u}_q + b_{u})),
\end{equation}
where $\mat{W}_G \in \mathbb{R}^{k \times d}$, is a mapping for shot representation $\mat{\Phi}_G$. We share $\mat{W}_{u}$ and $\mat{b}_{u}$ with the spatial attention model in Eq.~\ref{eq:SpatialH}. Similar to Eq.~\ref{eq:SpatialH}, in order to apply the summation $\oplus$, $\mat{u}_q$ is repeated $|G|$ times to have the same number of columns as $\mat{\Phi}_G$. Each column of matrix $\mat{\Psi}_G \in \mathbb{R}^{k \times |G|}$ is the combination of a single shot and the source sentence. 
An LSTM is employed to model the dynamics between shots and the source sentence:
\begin{equation}
    \mat{\Omega}^i_G = \textrm{LSTM}(\Psi_G^i) \quad (i=1...|G|),
\end{equation}
where $\mat{\Psi}_G^i$ is the $i$'th column of $\mat{\Psi}_G$. The output of this LSTM is a sequence of attention vectors corresponding to each shot $\mat{\Omega}_G = [\Omega_G^1, \Omega_G^2, ..., \Omega_G^{|G|}]$ (See Fig.~\ref{fig_temporal_attention}). However, we need to make probabilities out of these vectors and for this purpose we simply use a $softmax$ operator:
\begin{equation} \label{eq:temporalAttention}
    \mat{p}_{tp} = softmax(\mat{\Omega}_G^T \mat{w}_{tp}),
\end{equation}
where $\mat{w}_{tp} \in \mathbb{R}^{k \times 1}$ is a trainable weight vector and $\mat{p}_{tp} \in \mathbb{R}^{|G| \times 1}$ is the temporal attention of the shots and the final temporal-pooled representation is a weighted average over all the shot features $\mat{\Phi}_G$ with the attention model:
\begin{equation} \label{eq:teout}
\mat{u}_{tp} = \mat{\Phi}_G \mat{p}_{tp},
\end{equation}
where $\mat{u}_{tp} \in \mathbb{R}^{d}$ is of the same dimension as the spatial-pooled features and the source sentence representation. Its values are also bounded since $\mat{\Phi}_G$ is obtained by passing through a $\tanh(\cdot)$ activation in Eq. ~\ref{eq:upsilon}. By this temporal attention model, we capture the dynamics of the visual features which are related to the source sentence representation.

\subsection{Inference of the Missing Word}
Here, we discuss how to infer the missing word or fill in the blank. Let $\beta$ be the vocabulary to fill in the blank ($|\beta|$ as its size). We aim to find a probability for each word candidate in $\beta$, which needs a joint representation (summation fusion \cite{Yang2015}) of all three components as mentioned earlier:
\begin{equation}
    \mat{u} = [\mat{u}_q + \mat{u}_{sp} + \mat{u}_{tp}],
\end{equation}
where $\mat{u} \in \mathbb{R}^{d}$ is an joint representation of all three features: source sentence representations, spatially- and temporally-pooled visual representations. Regarding equations ~\ref{eq:qout}, ~\ref{eq:spout} and  ~\ref{eq:teout}, all three vectors $\mat{u}_q$, $\mat{u}_{sp}$ and $\mat{u}_{tp}$ are bounded, since they have been obtained by passing through a $\tanh(\cdot)$ activation. They also have the same dimension $d$ and this makes the summation fusion \cite{Yang2015} applicable and effective. For the final inference of the missing word, we compute a probability of each candidate word as follows:
\begin{equation}
    P_{blank} = softmax(\mat{W}_{blank} \mat{u}),
\end{equation}
where $\mat{W}_{blank} \in \mathbb{R}^{|\beta| \times d}$. It is followed by a multinomial logistic regression ``$softmax$'' to find the probabilities vector $\mat{P}_{blank} \in \mathbb{R}^{|\beta|}$. Based on Eq.~\ref{eq:theory}, the final answer is:
\begin{equation}
    \hat{b} = \argmax_{b \in \beta}{P_{blank}(b)}.
\end{equation}

\section{Experiments}
\label{sec_experiments}
We perform experiments on two datasets: the original LSMDC Dataset~\cite{maharaj2016dataset} to evaluate the single blank VFIB problem (i.e. there is only one blank in the source sentence), and an extended LSMDC Movie Dataset to evaluate our performance on the multiple blanks VFIB problem (i.e. there are multiple blanks in the sentence). 

\subsection{LSMDC Movie Dataset (Single Blank)}

In this set of experiments, we use the movie dataset ~\cite{maharaj2016dataset,rohrbach15cvpr, AtorabiM-VAD2015}, which has been used in Large Scale Movie Description and Understanding Challenge (LSMDC) ~\cite{maharaj2016dataset,Rohrbach2016}. Movies are a rich source of visual information and become much more valuable when proper textual meta-data is provided. Movies benefit from many textual data like the subtitle, audio descriptions and also movie synopsis. LSMDC dataset consists of respectively ``$91,908$'', ``$6,542$'', ``$10,053$'' and ``$9,578$'' movie clips as Training, Validation, Public and Private Test sets. We use the standard splits provided by~\cite{maharaj2016dataset, Rohrbach2016}. Each clip comes with a sentence annotated by an expert. There can be multiple source sentences built for one clip. We use respectively ``$296,960$'', ``$21,689$'' and  ``$30,349$'' samples as training, validation and test as the standard split provided by~\cite{Rohrbach2016}.

\subsubsection{Quantitative Results}
Here, we compare our proposed method with other approaches and baselines to show its superior performance for the VFIB task. We have chosen some of these baselines from methods for visual question answering problem, which are applicable to this problem as well. The comparison table (Table \ref{table_results}) has four parts. The first part is for methods which only use the text to find the missing word; the second part is for methods which just use the video; the third part is for methods which use both text and video; and the last part is for different configurations of the proposed method. We report the accuracy (same as in ~\cite{Rohrbach2016}) of each method which is the ratio of number of missing words that are inferred correctly to the total number of blanks.  Here are some details about these methods:

\begin{table}
\begin{center}
\begin{tabular}{|lc|}
\hline
\textbf{Method}  & \textbf{Accuracy} \\
\hline\hline
\textbf{Text Only} &  \\
Random Guess & 0.006 \\
LSTM  Left Sentence & 0.155 \\
LSTM  Right Sentence & 0.165 \\
BiLSTM      & 0.320 \\
\textbf{Our Sentence Encoding} & \textbf{0.367} \\
\textit{Human}~\cite{maharaj2016dataset} & \textit{0.302}\\
\hline
\textbf{Video Only} & \\
BiLSTM Just Video        & 0.055 \\
\hline
\textbf{Text +  Video} &  \\
GoogleNet-2D~\cite{maharaj2016dataset} & 0.349 \\
C3D~\cite{maharaj2016dataset} & 0.345 \\
GoogleNet-2D-Finetuned~\cite{maharaj2016dataset}  & 0.353\\
C3D-Finetuned~\cite{maharaj2016dataset} & 0.357 \\
Video + Textual Encoding~\cite{Ren2015} & 0.341\\
2Videos + Textual Encoding~\cite{Ren2015} & 0.350\\
Ask Your Neurons~\cite{malinowski2015ask} & 0.332\\
SNUVL~\cite{yu2016video} & 0.380 \\
SNUVL (Ensembled Model)~\cite{yu2016video} & 0.407 \\
\textit{Human}~\cite{maharaj2016dataset} & \textit{0.687} \\
\hline
\textbf{Ours} & \\
Single Model (w/o Spatial Attention) & 0.390 \\
Single Model (w/o Temporal Attention) & 0.392 \\
Single Model (w/o LR/RL LSTM) & 0.387 \\
\textbf{Single Model} & \textbf{0.409} \\
\textbf{Ensembled Model} & \textbf{0.434} \\
\hline
\end{tabular}
\end{center}
\caption{\label{table_results}Results on ``Movie Fill-in-the-Blank'' dataset.}
\end{table}

 \textbf{LSTM Left/Right Sentence} fills the blank by just looking at the left/right fragment of the missing word. This experiment shows that both fragments are equally important. \textbf{BiLSTM} finds the missing word based on a  BiLSTM ~\cite{schuster1997bidirectional}, which encodes the input sentence using two different LSTMs; one takes the input from the last word to the first word and the other one in the reverse. The blank word is recovered based on BiLSTM's output in missing word location. \textbf{Our Sentence Embedding} as described in section~\ref{seq:Qemb}. This approach finds the missing word by using just vector $\mat{u}_q$, without any visual features. For a fair comparison with BiLSTM method, we have fixed the LSTM cells sizes and also the word embedding lengths in all the experiments. \cite{maharaj2016dataset} provides a few baselines using \textbf{GoogleNet}~\cite{szegedy2015going} and \textbf{C3D}~\cite{tran2015learning} features. The difference between the baselines in\cite{maharaj2016dataset} and ours, shows the actual importance of our attention models and integration between the textual encoding and visual modules in our method. \textbf{Video+ Textual Encoding} corresponds to ``IMG+LSTM'' in ~\cite{Ren2015}. Key-frames are passed through the VGG-19 pre-trained network to extract $4,096$ dimensional vector of ``fc7'' layer for each key-frame. Then, a max-pooling over all the features of all frames will generate a video feature vector and the rest of steps are the same as explained in~\cite{Ren2015}. We have used simple BiLSTM instead of LSTM to deal with two fragments of sentence in VFIB.  \textbf{2Videos+ Textual Encoding}~\cite{Ren2015}, similar to previous case, uses two different representations of the video. One is attached to the beginning of each fragment and the other one to the end. \textbf{Ask Your Neurons}~\cite{malinowski2015ask} encodes the visual CNN feature and concatenate with each of words and pass them through left and right LSTMs one by one. The answer is inferred based on the last output of LSTMs. \textbf{SNUVL}~\cite{yu2016video} is the best reported method on LSMDC FIB. It uses a concept detection method over the videos, following by an attention model over the detected concepts, to find the missing word. \textbf{Ensemble model}~\cite{opitz1999popular} is a technique to boost the performance, when the optimization process reaches different local optimums, based on random factors like initialization. We train the model multiple times with different initializations and sum the final scores from all the trained models. We compare ensemble model results reported by~\cite{yu2016video} with ours.
 
We also test our model performance by removing each of components, namely spatial attention, temporal attention and also replacing our source sentence encoding with the BiLSTM in baselines. These experiments show that all the components contribute to our model and the model is not biased by just one component.

\subsection{Multiple Blanks VFIB}
\begin{figure}
\begin{center}
   \includegraphics[width=0.75 \linewidth]{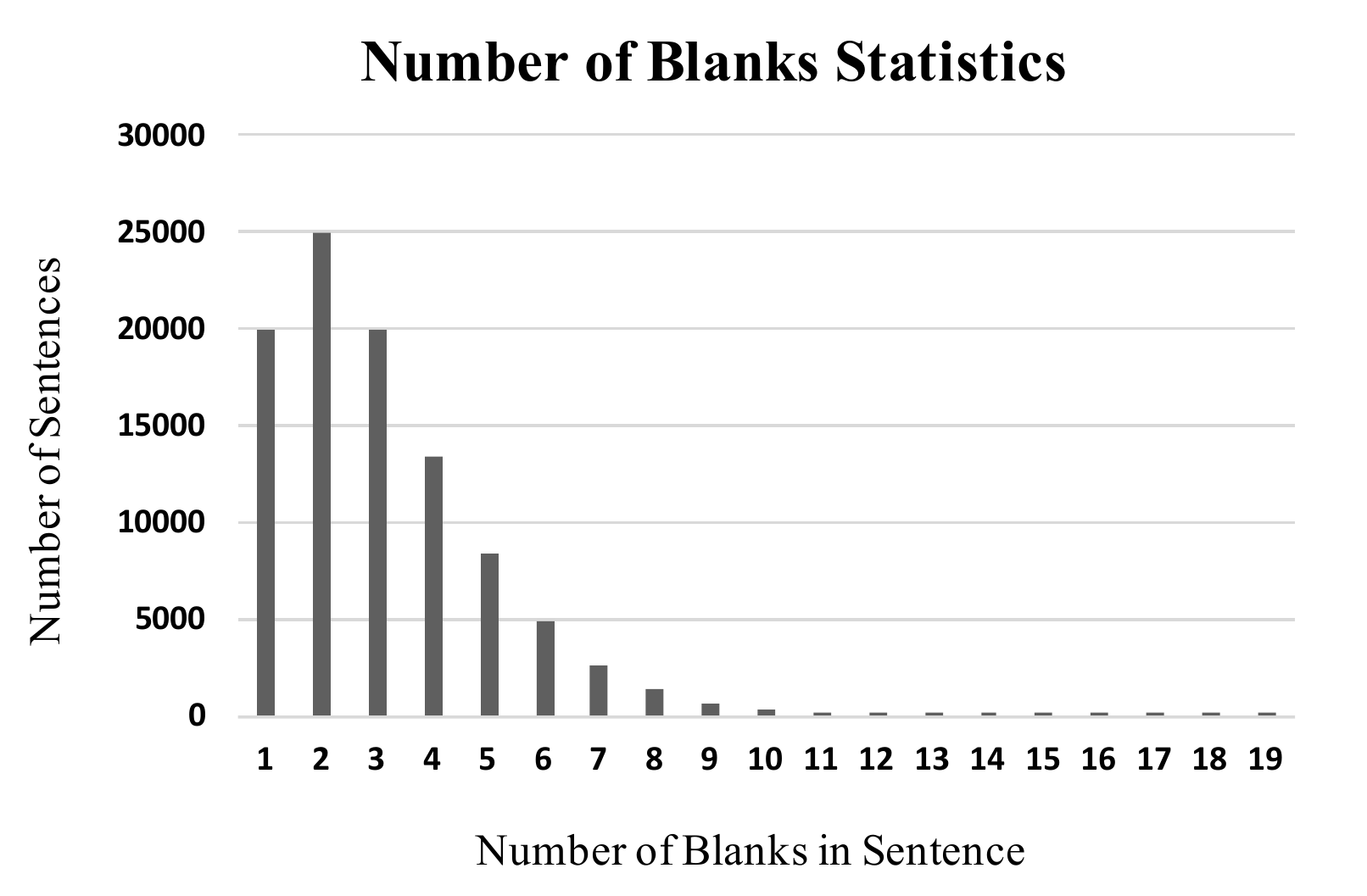}
\end{center}
   \caption{Number of sentences as a function of the  number of blanks in training set of LSMDC.}
\label{fig:BlankStats}
\end{figure}

In this section we explore a harder version of the VFIB problem where more than one word is missing from the sentence. We have generated a new dataset based on the original LSMDC Movie Dataset by inserting  multiple blanks in the sentences, and we call it the ``Extended LSMDC Movie Dataset''. To be specific, we remove all the words which has appeared at least once as a blank in the original LSMDC dataset from all the sentence. In this case, most of sentences have more than one blank and this makes the LSMDC dataset suitable to be extended for multiple blanks problem. In Figure~\ref{fig:BlankStats}, we show some statistics about the number of blanks in sentences. About $79.3\%$ of the sentences have more than one blank. To clarify, in each sentence, there are known number of blanks (with known locations), but there are various number of blanks in different sentences. For multiple blanks' experiments, we include all the sentences with one or more blanks in all sets and also for the evaluation, we consider equal value for all the blanks.

We employ two strategies to encode the source sentence for the multiple blanks VFIB problem. For the first one, we consider the left and right fragments of a missing word, as a fragment from that word to the next blank, or if there is not any other one, to the end of the sentence (both left and right fragments will be built). For example, for the source sentence \textit{``She took her \underline{Blank1} out of the garage and \underline{Blank2} at the house for a moment.''}. The left phrase of ``Blank1" is  \textit{``She took her''} and right phrase is \textit{``out of the garage and"} and we find the left and right phrases for ``Blank2" with the same approach as well. We call it the \textbf{``Subdivision"} approach since it makes multiple fragments out of the source sentence and each blank has one left and one right fragment. The second approach  is to remove all other blanks and treat them as left and right fragments as normal. In our example, the left fragment of ``Blank2'' is \textit{``She took her out of the garage''} and the right fragment of the ``Blank2'' is \textit{``out of the garage and at the house for a moment''}. In this case, we deal with each blank similar to single blank problem and we just ignore other blanks in each of left and right fragments. We call it \textbf{``Masking''} approach since we are masking the other missing words from each fragment. After finding left and right fragments based on any of these approaches, we can apply our method or any other baselines (Table~\ref{table_multiple_blank}).


\begin{table}
\begin{center}
\begin{tabular}{|lc|}
\hline
\textbf{Method}  & \textbf{Accuracy} \\
\hline
\hline
\textbf{Baselines} \\
Random Guess & 0.006 \\
2Videos + Textual (Subdivision)~\cite{Ren2015} & 0.136\\
2Videos + Textual (Masking)~\cite{Ren2015} & 0.177\\
\hline
\textbf{Ours} \\
Text Only (Subdivision) & 0.136 \\
Text + Video (Subdivision) & 0.148 \\
Text Only (Masking) & 0.173 \\
\textbf{Text + Video (Masking)} & \textbf{0.192} \\
\hline
\end{tabular}
\end{center}

\caption{Results on the LSMDC Dataset (Multiple Blanks). Our method has superior results.}
\label{table_multiple_blank}
\end{table}

\begin{figure*}
\begin{center}
   \includegraphics[width=0.90 \linewidth]{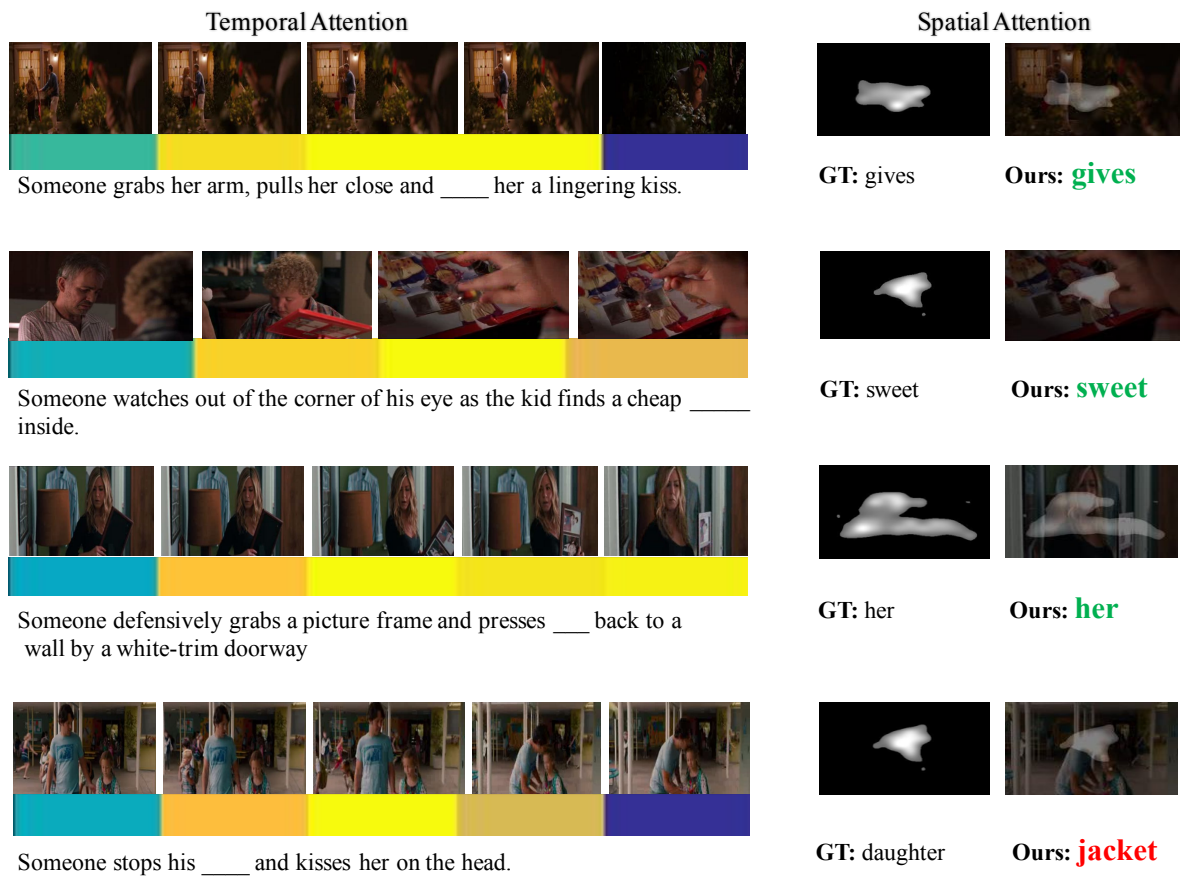}
\end{center}
   \caption{On the left we show representative frames from different shots. The colors below the frames show the temporal attention: yellow/blue means the  most/least attention. On the right, we show an spatial attention map obtained by our method and also we show the attention map on one of selected key-frames.}
\label{fig:Qualitative}
\end{figure*}

\subsection{Qualitative Results}
In Fig. ~\ref{fig:Qualitative}, we show some qualitative results. For generating the attention map, we have reshaped the  $\mat{p}_{sp} \in \mathbb{R}^{m=196}$ in Eq. ~\ref{eq:SpatialAttentionVector} into a $14 \times 14$ matrix, then  up-sampled it back to the original frames size. We smooth the attention map by a Gaussian filter and also  suppress low intensity pixels to be zero. Brighter parts  have higher attention score than the darker parts.
For the temporal attention model, we extract the $\mat{p}_{tp}$ vector as the temporal attention. In Fig. ~\ref{fig:Qualitative}, we show one frame from each shot and the color bar under the sequence of shots shows the attention scores. Yellow presents the maximum attention and blue is the minimum one. We provide more qualitative results in supplementary materials.

\subsection{Implementation Details} \label{featExtraction}
We have use VGG-19 ~\cite{Simonyan14c} network, pre-trained on ImageNet~\cite{deng2009imagenet}, and extract last pooling layer (``pool5'') as our spatial visual features consumed in section~\ref{SRep}. The output feature map is a $14 \times 14 \times 512$ matrix which can be reshaped as a $196 \times 512$ matrix and each of $512$ dimensional vectors are representing a $32 \times 32$ pixels region of input frame. We believe any other very deep CNN network like GoogLeNet~\cite{szegedy2015going} or ResNet~\cite{he2015deep} can produce similar results. We extract and pass the frames through this network with 2fps rate.
For temporal attention in section ~\ref{TeSe}, we use pre-trained 3D CNN (C3D) network~\cite{tran2015learning} pre-trained on~\cite{karpathy2014large} and followed settings defined in~\cite{tran2015learning}. We extract the ``fc6'' output of the network for each $16$ frames (one shot) of videos. We assume each video has $10$ shots. For shorter videos, we use all-zero vectors for remaining shots and for longer ones we uniformly select $10$ shots. We will include more details about the implementation in supplementary materials.
\section{Conclusion}
\label{sec_conclusion}
We proposed a new method for the Video-Fill-in-the-Blank (VFIB) problem which takes advantage from the ``source sentence'' structure and also spatial-temporal attention models. We have introduced ``external memory'' to deal with the complexity of ``source sentence'' in VFIB problem. All the required mathematical equations to make the model learn from multiple inputs (text, temporal/spatial features) are provided and discussed. We achieved superior performance over all other reported methods. Also, an extension and more general version of VFIB which deals with multiple blanks in sentences, has been introduced and discussed.

{\small
\bibliographystyle{ieeetr}
\bibliography{33}
}

\end{document}